\documentclass{article} 
\usepackage{nips10submit_e,times}
\usepackage{amsmath}
\usepackage[dvips]{graphicx}
\usepackage{subfigure}

\title{Sparse Group Restricted Boltzmann Machines}

\author{
Heng Luo \\
Department of Computer Science\\
Shanghai Jiao Tong University\\
\texttt{hengluo@sjtu.edu.cn} \\
\And
Ruimin Shen \\
Department of Computer Science\\
Shanghai Jiao Tong University\\
\texttt{rmshen@sjtu.edu.cn} \\
\AND
Changyong Niu \\
Department of Computer Science \\
Zhengzhou University \\
\texttt{cyniu@sjtu.edu.cn} \\
}

\nipsfinalcopy 

\begin{document}

\maketitle

\begin{abstract}
Since learning is typically very slow in Boltzmann machines, there
is a need to restrict connections within hidden layers. However, the
resulting states of hidden units exhibit statistical dependencies. Based on this observation, we propose
using $l_1/l_2$ regularization upon the activation possibilities of hidden units in restricted Boltzmann machines
to capture the loacal dependencies among hidden units. This regularization not only
encourages hidden units of many groups to be inactive given observed data but also
makes hidden units within a group compete with each other for modeling observed data.
Thus, the $l_1/l_2$ regularization on RBMs yields sparsity at both the group and the hidden unit levels. We call RBMs trained with the regularizer \emph{sparse group} RBMs. The proposed sparse group RBMs are applied
to three tasks: modeling patches of natural images, modeling
handwritten digits and pretaining a deep networks for a
classification task. Furthermore, we illustrate the regularizer can
also be applied to deep Boltzmann machines, which lead to sparse group
deep Boltzmann machines. When adapted to the MNIST data set, a
two-layer sparse group Boltzmann machine achieves an error rate of
$0.84\%$, which is, to our knowledge, the best published result on
the permutation-invariant version of the MNIST task.
\end{abstract}

\section{Introduction}

Restricted Boltzmann Machines (RBMs) [1,2] recently have become very popular
because of their excellent ability in unspervised learning, and have been
successfully applied in various application domains, such as dimensionality
reduction [3], Object Recognition [4] and others.

In order to obtain efficient and exact inference, there are no connections within the hidden layer in RBMs. But by considering statistical dependencies of states of hidden units we may learn a more powerful generative model [5] although directly learning horizontal connections in hidden layers leads to an inefficient inference. To consider, to some extent, statistical dependencies of states of hidden units and meanwhile keep the exact and efficient inference in RBMs, we introduce a $l_1/l_2$ regularizer on the activation possibilities of hidden units given training data.

$l_1/l_2$ regularizers have been intensively studied in both the statistics community [6] and machine learning community [7]. Usually the $l_1/l_2$ regularizer (or group lasso) only leads to sparsity at the group level but not within a group. Friedman et al [8] present a more general regularizer that blends $l1$ and $l_1/l_2$ regularizer. Using this regularizer, the linear model can yield sparsity both at the group level and within the group.

In this paper, we show that introducing the $l_1/l_2$ regularizer on the activation possibilities can yield sparsity at not only the group level but also the hidden unit level because of the Logistic differential equation. In the experiment section, we show that the sparse group RBM can be a better and sparser generative model than RBMs. And using the sparse group RBM can also learn more discriminative features.

\section{Restricted Bolzmann Machines and Contrastive Divergence}

A Restricted Boltzmann Machine is a two layer neural network with
one visible layer representing observed data and one hidden layer as
feature detectors. Connections only exist between the visible layer
and the hidden layer. Here we assume that both the visible
and hidden units of the RBM are binary. The models below can be
easily generalized to other types of units [9].
The energy function of a RBM is defined as
\begin{equation} \label{eq:1}
E(x,h)=-\sum_{i,j}x_{i}h_{j}w_{ij}-\sum_{i}x_{i}b_{i}-\sum_{j}h_{j}c_{j}
\end{equation}
where  $x_{i}$  and  $h_{j}$  denote the states of the $i$th
visible unit and  the $j$th  hidden unit, while $w_{ij}$
represents the strength of the connection between them. $b_{i}$
and $c_{j}$ are the biases of the visible and hidden units respectively.

Based on the energy function, we can define the joint distribution of $(x,h)$
\begin{equation} \label{eq:2}
P(x,h)=\frac{\exp (-E(x,h))}{Z}
\end{equation}
where $z$  is the partition function $Z=\sum_{x,h}\exp(-E(x,h))$.

The activation probabilities of the hidden units are
\begin{equation} \label{eq:3}
\begin{split}
P(h_{j}|x) &=sigmoid(x^{T}w_{.j}) \\
           &=\frac{1}{1+\exp(-x^{T}w_{.j})}
\end{split}
\end{equation}
where $w_{.j}$  denotes the  $j$th column of  $W$ , which is the
connection weights between the  $j$th  hidden unit and all
visible units. Because of  $x^{T}w_{.j}=\cos (\alpha)\|
w_{.j}\|_{2}\| x\|_{2}$  ($\alpha$ is the angle of the
vector $w_{.j}$ and $x$), the activation probability can be
interpreted as the similarity between $x$  and the feature $w_{.j}$
in data space.

The activation probabilities of the visible units are
\begin{equation} \label{eq:4}
P(x_{i}|h)=sigmoid(w_{i.}h)
\end{equation}

The marginal distribution over the visible units actually is a model
of products of experts [1,2]
\begin{equation} \label{eq:5}
\begin{split}
P(x) = \frac{\prod_{j}(1+\exp (x^{T}w_{.j}))}{Z}
\end{split}
\end{equation}
From Equation \ref{eq:5} we can deduce that each expert (hidden unit) will
contribute probabilities according to the similarity between its
feature and the data vector $x$. The feature of the $j$th expert
(the $j$th  hidden unit), $w_{.j}$ can be seen as a prototype
defined in data space [2].

The objective of generative training of a RBM is to model the
marginal distribution of the visible units $P(x)$. To do this, we
need to compute the gradient of the training data likelihood
\begin{equation} \label{eq:6}
\begin{split}
\frac{\partial\log P(x^{(n)})}{\partial\theta}=-<\frac{\partial
E(x^{(n)},h)}{\partial\theta}>_{P(h|x^{(n)})} + <\frac{\partial
E(x,h)}{\partial\theta}>_{P(x,h)}
\end{split}
\end{equation}
where $<.>_{P}$ is the expectation with respect to the distribution
$P$. The second term of Equation \ref{eq:6} is intractable since
sampling the distribution  $P(x,h)$  requires prolonged Gibbs
sampling. Hinton [1] shows that we can get very good
approximations to the second term when running the Gibbs sampler
only $k$ steps (the most commonly chosen value of $k$ is 1), initialized
from the training data. Named Contrastive Divergence (CD) [1],
the algorithm updates the feature of the $j$th hidden unit
seeing training data set $\{x^{(1)},x^{(2)},...,x^{(L)}\}$
\begin{equation} \label{eq:7}
\Delta w_{.j}=\frac{1}{L}\sum_{l=1}^{L}{P(h_{j}=1|x^{(l)}) \cdot x^{(l)}-P(h_{j}=1|x^{(l)-}) \cdot x^{(l)-}}
\end{equation}
where $x^{(l)-}$ is sampled from $P(x|h^{(l)})$
($h^{(l)}$ sampled from $P(h|x^{(l)})$). The first term of
Equation 7 will decrease energies of the RBM at $\{x^{(1)},x^{(2)},...,x^{(L)}\}$.
The second term will increase energies at $\{x^{(1)-},x^{(2)-},...,x^{(L)-}\}$ which possible are
near the training data set and have low energies in the RBM. Iteratively updating
parameters ensures that the probability distribution defined by the energy function has the desired local structure.
Furthermore from Equation \ref{eq:7}, one specific hidden unit will be responsible for decreasing (or increasing) only a subset of training data (or negative sample), which is selected by the vector of activation probabilities, $(P(h_{j}=1|x^{(1)}),...,P(h_{j}=1|x^{(L)}))^{T}$ (or $(P(h_{j}=1|x^{(1)-}),...,P(h_{j}=1|x^{(L)-}))^{T}$). Thus the learning algorithm ensures that different hidden units learn different features.

\section{Sparse group RBMs}

As discussed in Section 1, directly learning the statistical dependencies
between hidden units is inefficient. To alleviate this problem, firstly we
averagely divide hidden units into predefined non-overlapping groups to restrain the dependencies within these groups.
Secondly, instead of directly learning the dependencies we penalize the overall activation level of a group to force hidden units in the group to compete with each other. To implement the two above intuitions we introduce a mixed-norm ($l_1/l_2$) regularizer on the activation possibilities of hidden units given the training data.

Assuming a RBM has $F$ hidden units, let $\mathcal{H}$ denote the set of all hidden units' indices:
$\mathcal{H}=\{1,2,...,F\}$. The $k$th group is denoted by $\mathcal{G}_{k}$ where $\mathcal{G}_{k}\subset\mathcal{H},k=1,...,K$. Given a grouping $\mathcal{G}$ and a data point $x^{(n)}$, the $k$th group norm $N_k$ is given by
\begin{equation} \label{eq:8}
N_k=\sqrt{\sum_{m\in G_{k}}P(h_{m}=1|x^{(n)})^2}
\end{equation}
which is the Euclidean ($l_2$) norm of the vector composed of these activation possibilities and considered as the overall activation level of $k$th group. Given the group norms, the mixed-norm is
\begin{equation} \label{eq:9}
\sum_{k=1}^{K}|N_k|=\sum_{k=1}^{K}\sqrt{\sum_{m\in G_{k}}P(h_{m}=1|x^{(n)})^2}
\end{equation}
which is the $l_1$ norm of the vector composed of the group norms.

We add the $l_1/l_2$ regularizer to the log-likelihood of training data (see Equation \ref{eq:6}). Thus, given training data $\{x^{(1)},x^{(2)},...,x^{(L)}\}$, we need to solve the following optimization problem
\begin{equation} \label{eq:9}
maximize_{W,b,c}\sum_{l=1}^{L}\log{P(x^{(l)})}-\lambda\sum_{k=1}^{K}\sqrt{\sum_{j\in \mathcal{G}_{k}}P(h_{j}=1|x^{(l)})^2}
\end{equation}
where $\mathcal{G}_{k}$ is the index set belonging to the $k$th group of hidden units
and $\lambda$ is a regularization constant.

\begin{figure}[h]
  \centering
  \subfigure[]{
    \label{fig:1:a} 
    \includegraphics[width=0.4\linewidth]{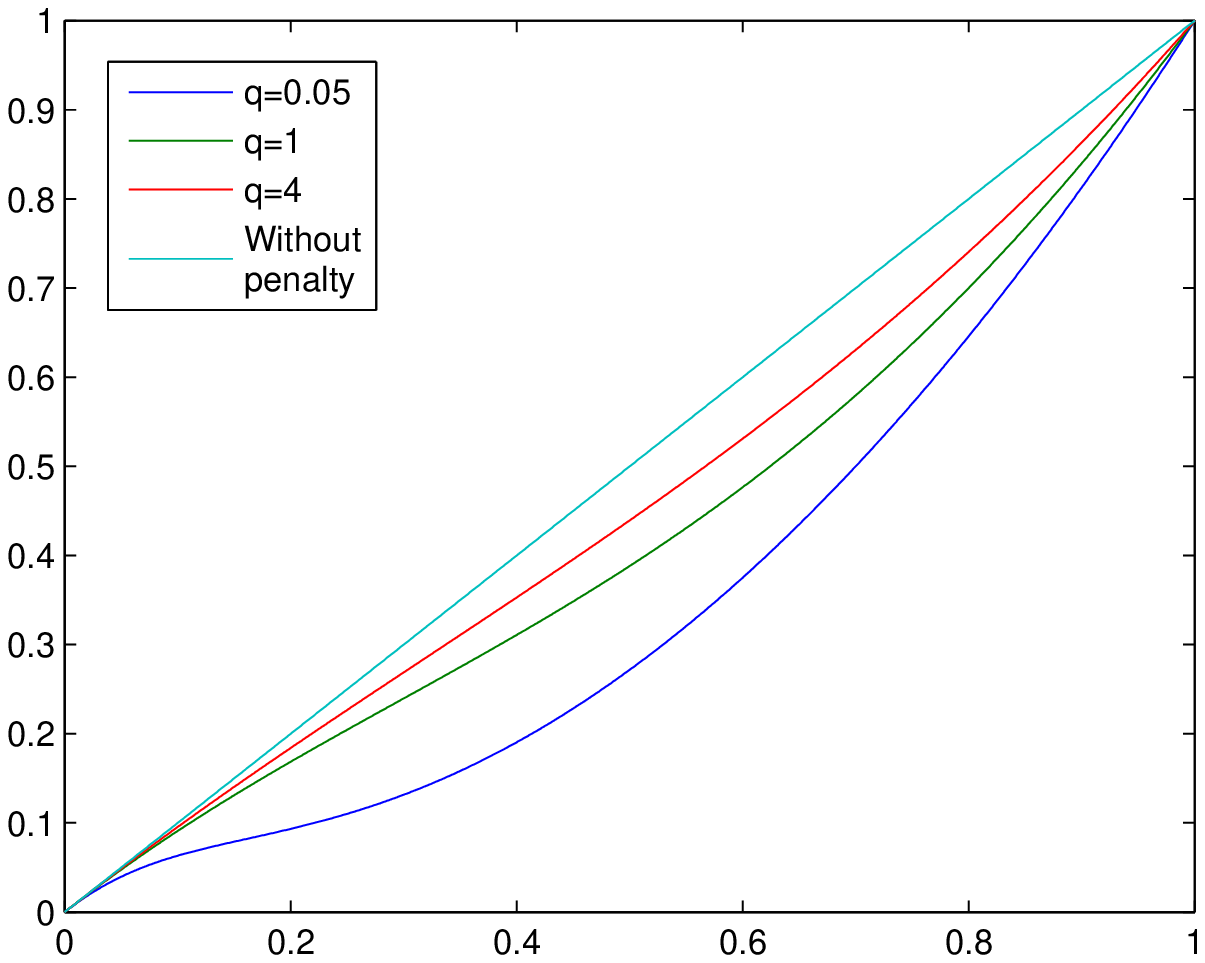}}
  \hspace{0.2in}
  \subfigure[]{
    \label{fig:1:b} 
    \includegraphics[width=0.4\linewidth]{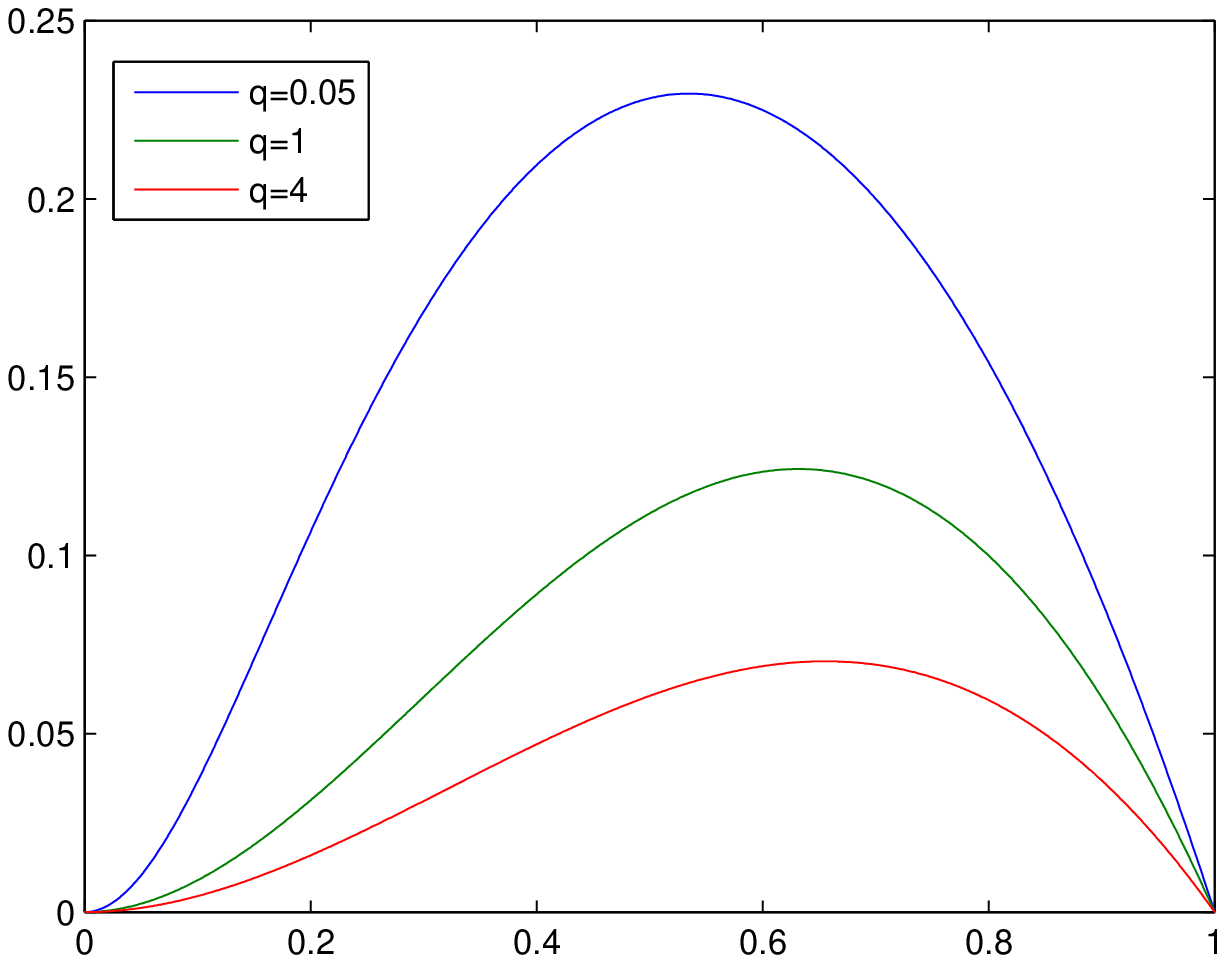}}
  \caption{Assuming the $j$th hidden unit belongs to the $k$th group and $q=\sum_{m\in G_{k},m \neq j}P(h_{m}=1|x^{(l)})^2$. The regularization constant $\lambda$, is set to 1. In both of two figures the horizontal axis represents activations of $j$th hidden unit. (a) Vertical axis: the coefficients of data $x^{(n)}$;
  (b) Vertical axis: the values of the second term in the coefficients of data $x^{(n)}$}
  \label{fig:6} 
\end{figure}

The effects of a $l_1/l_2$ regularization can be interpreted on two levels:
an across-group and a within-group level. On the across-group level, the group norms
$N_k$ behave as if they were penalized by a $l_1$ norm. In consequence, given observed data
some group norms are zero, which means the activation possibilities of all hidden units
in these groups are zero since the activation possibilities are non-negative.
On the within-group level, the $l_2$ norm will equally penalize the activation possibilities
of all hidden units in the same group. In other words, the $l_2$ norm does not yield sparsity within the group. However, when applied the $l_1/l_2$ on the activation possibilities it is an entirely different story because of the Logistic differential equation. Below we will discuss it in detail.

By introducing this regularizer the Equation \ref{eq:7} is changed to the following equation
\begin{equation} \label{eq:10}
\frac{1}{L}\sum_{l=1}^{L}{\Big(P(h_{j}=1|x^{(l)})-\lambda\frac{P(h_{j}=1|x^{(l)})^2P(h_{j}=0|x^{(l)})}{\sqrt{\sum_{m\in G_{k}}P(h_{m}=1|x^{(l)})^2}} \Big) \cdot x^{(l)} - P(h_{j}=1|x^{(l)-}) \cdot x^{(l)-}}
\end{equation}
where $j\in G_{k}$. As discussion in Section 2, the decrease of the RBM's energies at the training data $x^{(l)}$ is determined by $P(h_{j}=1|x^{l})$ in Equation \ref{eq:7}. With the $l_1/l_2$ regularization the decrease
is determined by not only $P(h_{j}=1|x^{l})$ but also other hidden units' activation possibilities
in the same group. To interpret properties of this regularizer, we visualize the coefficients of data point $x^{(l)}$ in Figure \ref{fig:1:a} and the second term of the coefficients in Figure \ref{fig:1:b}.

$q=\sum_{m\in G_{k},m \neq j}P(h_{m}=1|x^{(l)})^2$ can be interpreted as a overall activation level of the hidden units in $k$th group except $j$th hidden unit. A small value of $q$ indicates that most of hidden units in the group are inactive for the data $x^{(l)}$. In Figure \ref{fig:1:a}, a smaller $q$ slow down the decrease of the RBM's energies at data point $x^{(l)}$. In consequence a hidden unit is penalized strongly if the activation possibilities of the remaining hidden units in the corresponding group are very small. Thus, the first property of the $l_1/l_2$ regularizer is that it encourages few groups to be active given observed data. This property yield the sparsity at the group level. \newline

In Figure \ref{fig:1:b}, the effect of the regularizer will vanish when $P(h_{j}=1|x^{(l)})$ is close to 0 or 1 because of the factor $P(h_{j}=1|x^{(l)})^2P(h_{j}=0|x^{(l)})$ in Equation \ref{eq:10}. A bigger $q$ indicates that most of hidden units in the group are active for the data $x^{(l)}$. In the Figure \ref{fig:1:b}, it can also be seen that the effects of the regularizer become smaller when the activation possibilities are close to 0 instead of them close to 1 because of the square of $P(h_{j}=1|x^{(l)})$. This can be interpreted as that hidden units in a group compete with each other for modeling data $x^{(l)}$. Only few of hidden units in a group will win this competition. Thus, the second property of the regularizer is that is results in only few hidden units to be active in a group. This property yields the sparsity within the group. Based on these two properties, we call RBMs trained by Equation \ref{eq:9} sparse group RBMs.

\section{Relationship to third-order RBMs}

A third-order RBM described in [10] is a mixture model whose components are RBMs. To a certain extent, a trained third-RBM defines a special \textbf{group sparse} representation for training data. Discussing the relationships between a third-RBM and sparse group RBMs will give us additional insights about the effects of $l_1/l_2$ regularizer for RBMs.

The energy function of a third-order Boltzmann machine is
\begin{equation} \label{eq:11}
E(x,h,z)=-\sum_{i,j,k}x_{i}h_{j}^{k}w_{ij}^{k}z_{k}
\end{equation}
where $z$ is a $K$-dimensional binary vector with $1$-of-$K$ activation and represents the cluster label. The responsibility of the $k$th component RBM is
\begin{equation} \label{eq:12}
\begin{split}
P(z_{k}=1|x) &= \frac{\sum_{h}P(x,z_{k}=1,h)}{\sum_{l=1}^{K}P(x,z_{l}=1)}
\\           &= \frac{\prod_{j}(1+\exp (x^{T}w_{.j}^k))}{\sum_{l=1}^{K} \prod_{j'}(1+\exp (x^{T}w_{.j'}^{l}))}
\\           &= \frac{\prod_{j}(1/(1-P(h_{j}^{k}=1|x))}{\sum_{l=1}^{K} \prod_{j'}(1/(1-P(h_{j'}^{l}=1|x))}
\end{split}
\end{equation}

A 3-order RBM with $K$ components can be seen as a regular RBM in which hidden units are divided into $K$ non-overlapping groups. Given a data $x$ the responsibility, $P(z|x)$ is used to pick one group's hidden units to respond to the data. In other words, data will be represented by only one group' hidden units and the states of hidden units in other groups will be set to $0$. From this perspective, a third-order RBM yields a special group sparsity given the training data.

The group (component) which has the bigger value of $\prod_{j}(1/(1-P(h_{j}^{k}=1|x))$ will more likely be responsible for the data $x$. Given the data $x$ the product $\prod_{j}(1/(1-P(h_{j}^{k}=1|x))$ can be interpreted as a measure of overall activation level of hidden units in the group. If more hidden units in the group are active, the overall activation level of the group is higher. However the products are unbounded and at very different numerical scales since any hidden unit's activation possibility ($P(h_{j}^{k}=1|x)$) in a group that is close to $1$ will make the product extremely big. To alleviate this problem, Nair and Hinton [10] introduced a temperature parameter $T$ to reduce scale differences in the products.

There are two major differences between 3-order RBMs and sparse group RBMs. Firstly, sparse group RBMs define a different overall activation level of a group's hidden units, which is the euclidean norm of the vector, $(P(h_j=1|x))_{j \in \mathcal{G}_{k}}$. Since this measure is bounded and in the interval $[0, |\mathcal{G}_{k}|]$, it can be avoided that one group with a too high overall activation level shields all of other groups. Secondly, as discussed in Section 3, sparse group RBMs yields sparsity at both the group level and the hidden unit level by regularization. It is a more flexible method than as done with third-order RBMs. Third-order RBMs directly divide training data into subsets by $P(z|x)$, where each subsets is modeled by hidden units in one specific group.

\section{Sparse group deep Bolzmann machines}

Salakhutdinov and Hinton [11] presented a learning algorithms for tractable training a deep multilayer Bolzmann machines, in which, unlike deep belief networks, hidden units will receive top-down feedback. Based on their learning algorithm, we illustrate that the regularizer can also be added to a deep Bolzmann machine. This leads to a sparse group deep Bolzmann machine. Taking a two-layer Boltzmann machine for example, the energy function is
\begin{equation} \label{eq:13}
E(x,h^{1},h^{2})=-x^{T}W^{1}h^{1}-x^{T}W^{2}h^{2}
\end{equation}

For training a sparse group deep Bolzmann machine, we propose the following optimization problem
\begin{equation} \label{eq:14}
maximize_{W^{1},W^{2}}\sum_{l=1}^{L}\log{P(x^{(l)})}-\lambda\sum_{k=1}^{K}\sqrt{\sum_{j\in \mathcal{G}_{k}^{1}}P(h_{j}^{1}=1|x^{(l)})^2}-\lambda^{'}\sum_{k=1}^{K'}\sqrt{\sum_{j\in \mathcal{G}_{k}^{2}}P(h_{j}^{2}=1|x^{(l)})^2}
\end{equation}
Given observed data the two activation probabilities can not be computed efficiently. Thus, following [11], we adopted the mean-field approximations for these two probabilities.

\section{Experiments}

We firstly applied sparse group RBMs to three tasks: modeling patches of natural images, modeling handwritten digits and pretaining a multilayer feedforward network for handwritten digits recognition. The first two tasks are adopted to evaluate the performances of sparse group RBMs (as a generative model) on modeling different types (real-valued and binary) of observed data. The third task partially accesses the performances of using features learned by sparse group RBMs on a discriminative task. We also trained a sparse group deep Boltzmann machine for handwritten digits recognition. \\
\\
A bigger group easily leads to a bigger $q$ (see Section 3), which make hidden units in this group receive weaker penalties and moderate the competitions among hidden units in this group. In the meanwhile using a bigger $\lambda$ can keep the competitions intense though $q$ is big. However, a big $\lambda$ may lead to a negative coefficients of $x^{(l)}$ in Equation \ref{eq:10} and prevent hidden units from modeling $x^{(l)}$ when $q$ is small. So the group size needs to be set small (empirically below 10). And in all experiments, the sparse group RBMs' parameter $\lambda$, is empirically set to $0.1$ which ensures the regularizer not to dominate the learning.

\subsection{Modeling patches of natural images}

Using regular RBMs trained on patches of natural images will learn relatively diffuse, unlocalized features. Lee et al. [12] proposed sparse RBMs to model natural images. Because sparse group RBMs yields sparsity at the hidden units level, we show sparse group RBMs can also be used for modeling patches of natural images.

The training data used consists of $100,000$ $14\times14$ patches randomly extracted from a standard set of $10$
$512\times512$ whitened images as in [13]. We divided all patches into mini-batches, each of which contained 200 patches, and updated the weights after each mini-batch.
\\
\begin{figure}[h]
\begin{center}
\includegraphics[width=0.8\linewidth]{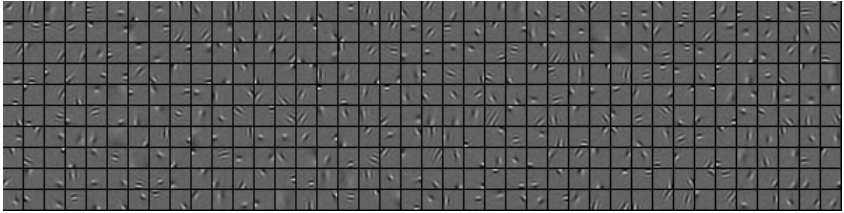}
\end{center}
\caption{Learned features with sparse group RBM trained on patches of natural images}
\label{fig:2}
\end{figure}

\begin{figure}[h]
\begin{center}
\includegraphics[width=0.8\linewidth]{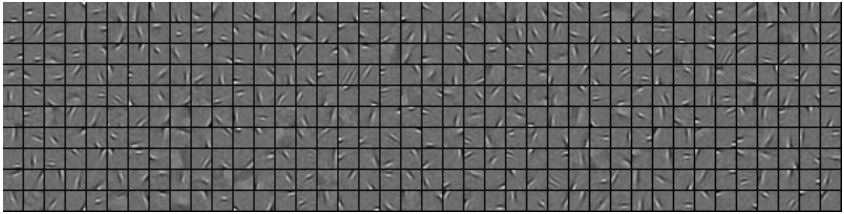}
\end{center}
\caption{Learned features with sparse RBM trained on patches of natural images}%
\label{fig:3}
\end{figure}

We trained a sparse group RBM with $196$ real-valued visible units and $400$ hidden units which are divided into $80$ uniform non-overlapping groups. There are 5 hidden units in each group. The learned features are shown in Figure \ref{fig:2}. The features are localized, oriented, gabor-like filters.

For comparison, we also trained a sparse RBM [12] with $400$ hidden units. The learned features are shown in Figure \ref{fig:3}. The sparse RBMs' parameters, $p$ and $\lambda$ are set to 50 and 0.02 as suggested in [12]. As discussed in Section 3, hidden units in a group compete with each other to model pathes, each hidden unit in the sparse group RBM is focused on modeling more subtle patterns contained in training data. As a result, the features learned with the sparse group RBM are much more localized than those learned with the sparse RBM.

\subsection{Modeling handwritten digits}

\begin{figure}[h]
\begin{center}
\includegraphics[width=0.8\linewidth]{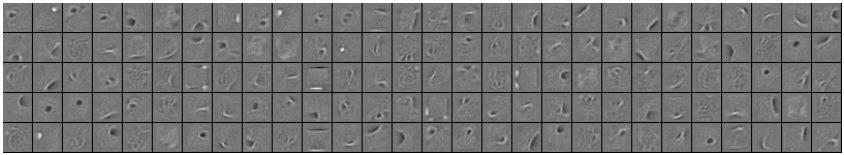}
\end{center}
\caption{Learned features with RBM trained on MNIST dataset}
\label{fig:4}
\end{figure}

We also applied sparse group RBMs algorithm to the MNIST handwritten digit dataset\footnote{http://yann.lecun.com/exdb/mnist/}. The training data, $60,000$ $28\times28$ images were divided into mini-batches, each of which contained 100 images. We trained two sparse group RBMs with $600$ hidden units and different group size ($3$ and $5$). We compare these models to a regular RBM with $600$ hidden units. The learned features are shown in Figure \ref{fig:4}. All of three models are trained by CD-1 for 50 epochs with a same configuration of learning rate, weight decay and momentum.

\begin{figure}[h]
\begin{center}
\includegraphics[width=0.8\linewidth]{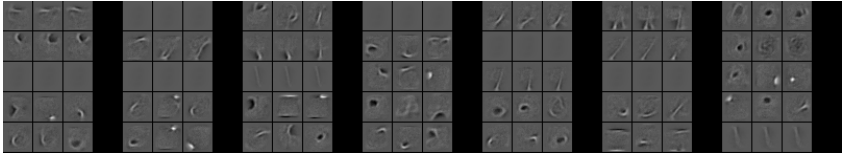}
\end{center}
\caption{Learned features with sparse group RBM trained on MNIST dataset. 3 patches in one row of every column are  visualizations of features of hidden units in one specific group.}
\label{fig:5}
\end{figure}

Due to space reasons, we show only the features of the sparse group RBM with group size 3 in Figure \ref{fig:5}. The results of the sparse group RBM with group size 5 are similar. Many features in Figure \ref{fig:5} look like different strokes of handwritten digits.

Although computing the exact partition function of a RBM is intractable, Salakhutdinov and Murray [14] proposed an Annealed Importance Sampling based algorithm to tractably approximate the partition function of an RBM. Using their method, the estimates of the lower bound on the average test log-probability are $-123$ for the regular RBM, $-104$ for the sparse group RBM with group size 3 and $-111$ for the sparse group RBM with group size 5. It can be seen that by adopting a proper regularization we can learn a better generative model on MNIST dataset than those without the regularization.

\begin{figure}[h]
  \centering
  \subfigure[]{
    \label{fig:6:a} 
    \includegraphics[width=0.4\linewidth]{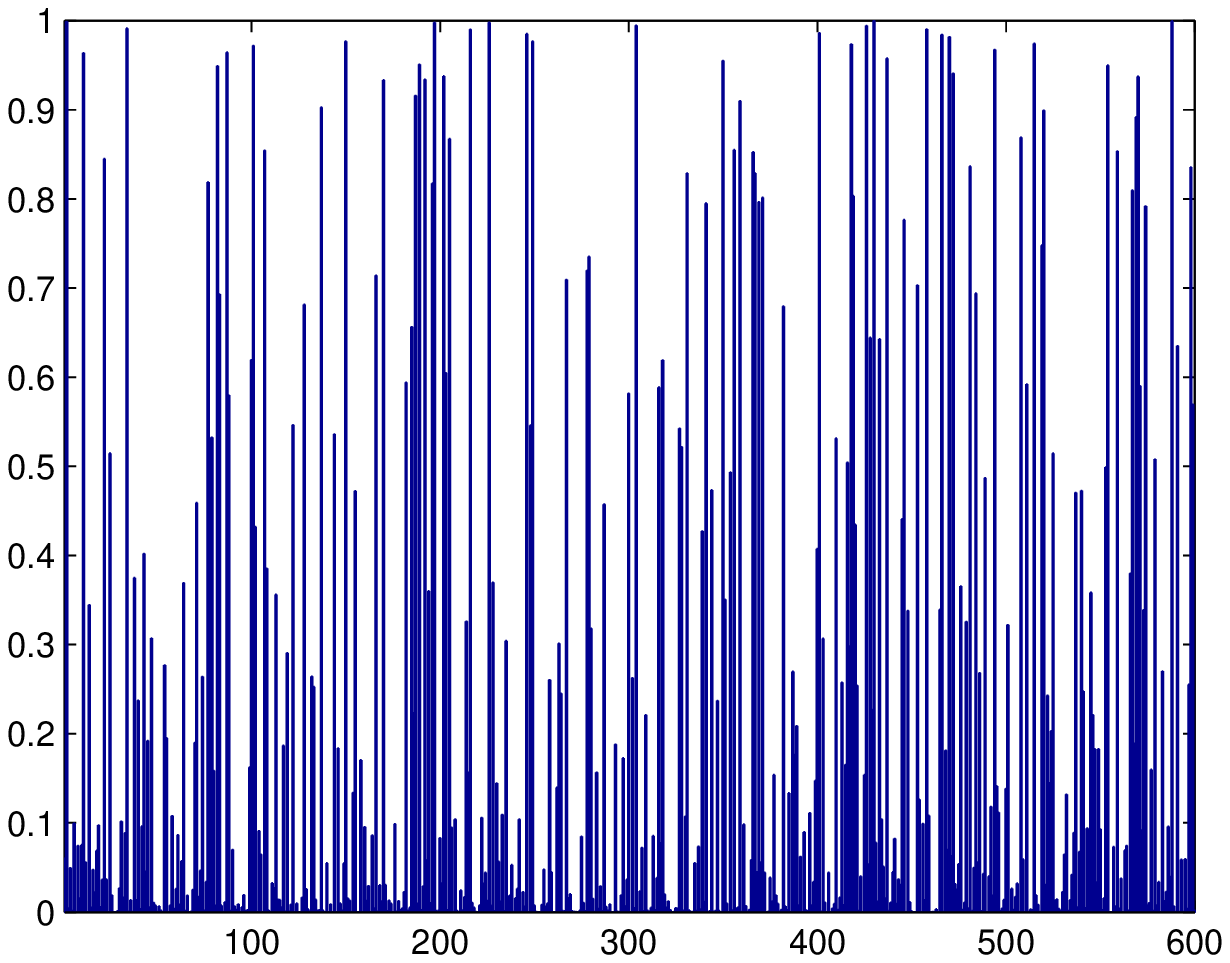}}
  \hspace{0.2in}
  \subfigure[]{
    \label{fig:6:b} 
    \includegraphics[width=0.4\linewidth]{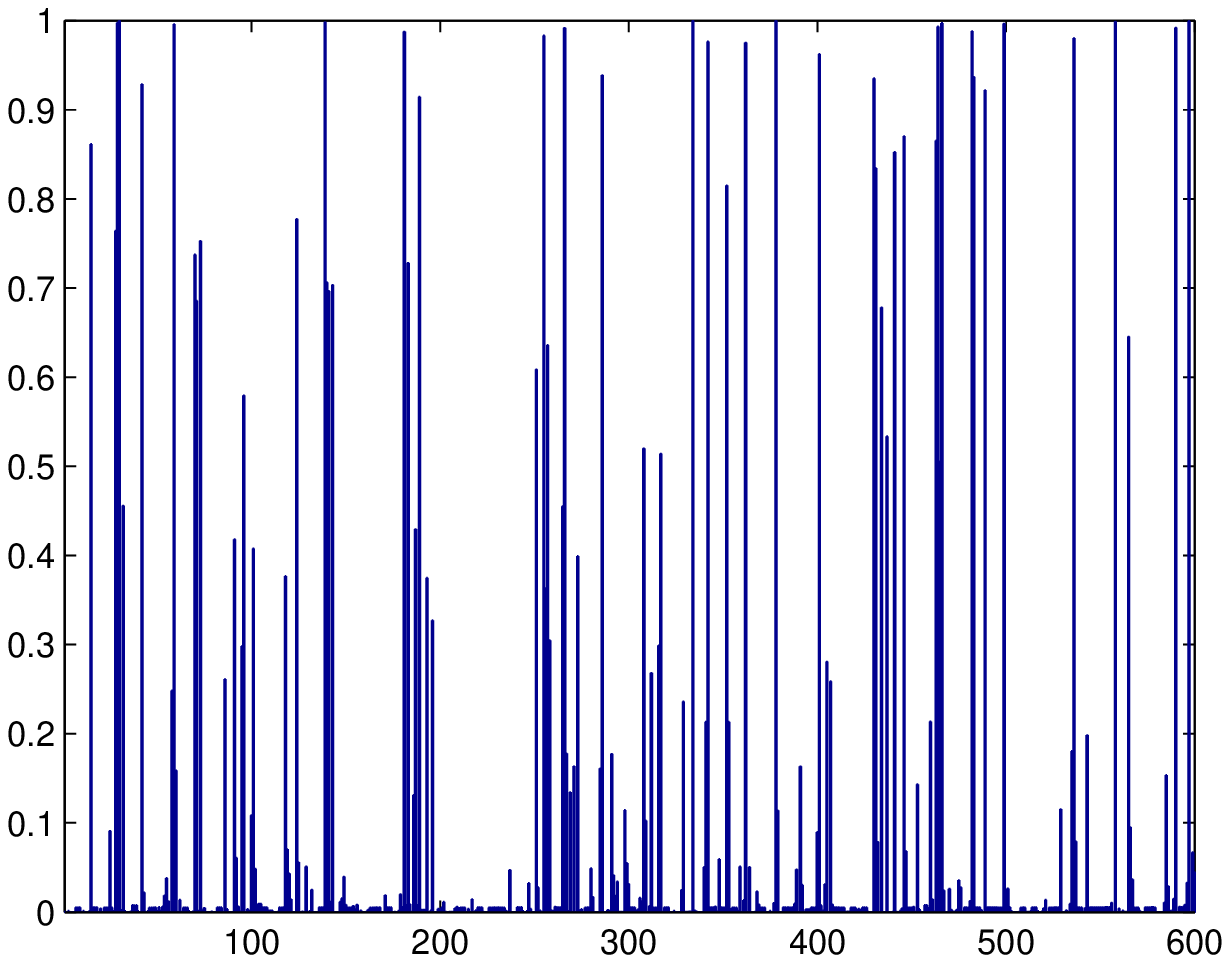}}
  \caption{(a) Activation possibilities computed under the regular RBMs. The sparseness of the vector is $0.49$; (b) Activation possibilities computed under the sparse group RBMs with group size 3. The sparseness is $0.68$.}
  \label{fig:6} 
\end{figure}

We use Hoyer's sparseness measure [15] to figure out how sparse representations learned by RBMs and sparse group RBMs. This sparseness measure evaluate the sparseness of a $D$ dimensional vector $v$ in the following way
\begin{equation} \label{eq:15}
sparseness(v)= \frac{\sqrt{D}-\frac{\sum_{i=1}^D |v_i|}{\sqrt{\sum_{i=1}^D v_{i}^{2}}}}{\sqrt{D}-1}
\end{equation}
This measure has good properties, which is in the interval $[0,1]$ and on a normalized scale. Its value more close to $1$ means that there are more zero components in the vector $v$. With every trained models, we can compute activation possibilities of hidden units given $10,000$ test images. Given any trained model this leads to new representations ($10,000$ $600$-dimensional vectors) of test data. The sparseness measures of the representations under the RBMs are in the interval $[0.36, 0.6]$, with an average of $0.5$. The sparseness measures of the representations under the sparse group RBMs with group size 3 and 5 are in [0.55, 0.75] and [0.51, 0.72]. The averages are $0.68$ and $0.65$, respectively. It can be seen that the sparse group RBMs can learn much more sparser representations than regular RBMs on MNIST data sset. Figure \ref{fig:6:a} visualizes the activation possibilities of hidden units, which are computed under the regular RBMs given an image from test set. Given the same image the activation possibilities computed under the sparse group RBMs are shown in Figure \ref{fig:6:b}.

\subsection{Using sparse group RBMs to pretrain deep networks}

One of the most important applications of RBMs is to use RBMs as building blocks layer-by-layer to pretrains greedily deep supervised feedforward neural networks [3]. We show that sparse group RBMs can also be used to initialize deep networks and achieve better performances of classification on MNIST dataset.

We use sparse group RBMs with different group size ($3$, $5$ and $10$) on MNIST dataset to pretain a 784-600-600-2100 networks. After pretraining, the multilayer networks are fine-tuned for 30 iterations using Conjugate Gradient. The networks initialized by the sparse group RBMs with group size, $3$, $5$ and $10$ achieve the error rates of $0.89\%$, $0.91\%$ and $0.99\%$, respectively. Using regular RBMs pretraining a 784-500-500-2000 network achieved the error rate of $1.14\%$ [3]. A network with the same architecture initialized by sparse RBMs gave a much worse error rate of $1.81\%$ [16].

\subsection{Sparse group deep Boltzmann machines}

We also trained a two layer ($500$ and $1000$ hidden units) sparse group Boltzmann machine on MNIST dataset. The group size is set to 10 for both of two layers. Firstly, we use two sparse group RBMs (784-500 and 500-1000) to initialize the deep network. Then the learning algorithm described in Section $5$ trains the sparse group deep Boltzmann machine. Finally we discriminative fine-tuned the network. The sparse group deep Bolzmann machine achieves the error rates of $0.84\%$ on the test set, which is, to our knowledge, the best published result on the permutation-invariant version of the MNIST task. The deep Boltzmann machine with the same architecture without the sparse group regularization resulted in the error rate of $0.95\%$ [11].

\section{Conclusions}

In this paper, we introduce $l_1/l_2$ regularization on the activation possibilities of hidden units in restricted Boltzmann machines. This leads to a better and sparser generative model, sparse group RBM.

\subsubsection*{Acknowledgments}

\subsubsection*{References}

\small{
[1] Hinton, G.E.: Training products of experts by minimizing contrastive divergence. {\it Neural Computation}, 14(8):1711-1800 (2002)

[2] Freund, Y., Haussler, D.: Unsupervised learning of distributions of binary vectors using 2-layer networks. In {\it Advances in Neural Information Processing Systems}, volume 4, pages 912-919 (1992)

[3] Hinton, G.E., Salakhutdinov, R.R.: Reducing the dimensionality of data with neural networks. {\it Science}, 313:504-507 (2006)

[4] Lee, H., Grosse, R., Ranganath, R., Ng, A.Y: Convolutional deep belief networks for scalable unsupervised
learning of hierarchical representations. In {\it Proceedings of the Twenty-sixth International Conference on Machine Learning}. ACM, Montreal (Qc), Canada. (2009)

[5] Garrigues, P., Olshausen, B.: Learning horizontal connections in a sparse coding model of natural images. {\it Advances in Neural Information Processing Systems}, vol. 20, pp. 505-512, (2008).

[6] Yuan, M., Lin, Y.: Model selection and estimation in regression
with grouped variables. {\it Journal of the Royal Statistical Society}, Series B 68(1), 49-67. (2007)

[7] Bach, F.: Consistency of the group Lasso and multiple kernel learning. {\it Journal of Machine Learning
Research}, 9:1179-1225, (2008).

[8] Friedman, J., Hastie, T., Tibshirani, R: A note on the group lasso
and the sparse group lasso, Technical report, Statistics Department,
Stanford University. (2010).

[9] Welling, M., Rosen-Zvi, M., Hinton, G.E.: Exponential family harmoniums with an application to information retrieval. In {\it Advances in Neural Information Processing Systems 17}, MIT Press (2005)

[10] Nair, V., Hinton, G.E.: Implicit Mixtures of Restricted Boltzmann Machines. In {\it Advances in Neural Information Processing Systems 21}, MIT Press (2009)

[11] Salakhutdinov, R.R., Hinton, G.E.: Deep Boltzmann machines. In {\it Proceedings of The Twelfth
International Conference on Artificial Intelligence and Statistics}, Vol. 5, pp. 448-455 (2009)

[12] Lee, H., Ekanadham, C., Ng, A.: Sparse deep belief net model for visual area V2. In {\it Advances in Neural Information Processing Systems 20 (NIPS¡¯07)}. MIT Press, Cambridge, MA. (2008)

[13] Olshausen, B.A., Field, D.J: Emergence of simple-cell receptive field properties by learning a sparse
code for natural images. {\it Nature}, 381:607-609. (1996)

[14] Salakhutdinov, R., Murray, I: On the quantitative analysis of deep belief networks. In {\it Proceedings of the Twenty-fifth International Conference on Machine Learning (ICML¡¯08)}, Vol. 25, pp. 872-879. ACM. (2008)

[15] Hoyer, P.O.: Non-negative matrix factorization with sparseness constraints, {\it Journal of Machine Learning Research}, 1457-1469. (2004)

[16] Swersky, K., Chen, B., Marlin, B., Freitas, N.: A Tutorial on Stochastic Approximation Algorithms for Training Restricted Boltzmann Machines and Deep Belief Nets. Information Theory and Applications (ITA) Workshop. (2010)

\end{document}